# HYPERNETWORKS

**David Ha**,* **Andrew Dai, Quoc V. Le**
Google Brain
`{hadavid,adai,qvl}@google.com`

## ABSTRACT

This work explores hypernetworks: an approach of using a one network, also known as a hypernetwork, to generate the weights for another network. Hypernetworks provide an abstraction that is similar to what is found in nature: the relationship between a genotype – the hypernetwork – and a phenotype – the main network. Though they are also reminiscent of HyperNEAT in evolution, our hypernetworks are trained end-to-end with backpropagation and thus are usually faster. The focus of this work is to make hypernetworks useful for deep convolutional networks and long recurrent networks, where hypernetworks can be viewed as relaxed form of weight-sharing across layers. Our main result is that hypernetworks can generate non-shared weights for LSTM and achieve near state-of-the-art results on a variety of sequence modelling tasks including character-level language modelling, handwriting generation and neural machine translation, challenging the weight-sharing paradigm for recurrent networks. Our results also show that hypernetworks applied to convolutional networks still achieve respectable results for image recognition tasks compared to state-of-the-art baseline models while requiring fewer learnable parameters.

## 1 INTRODUCTION

In this work, we consider an approach of using a small network (called a "hypernetwork") to generate the weights for a larger network (called a main network). The behavior of the main network is the same with any usual neural network: it learns to map some raw inputs to their desired targets; whereas the hypernetwork takes a set of inputs that contain information about the structure of the weights and generates the weight for that layer (see Figure 1).

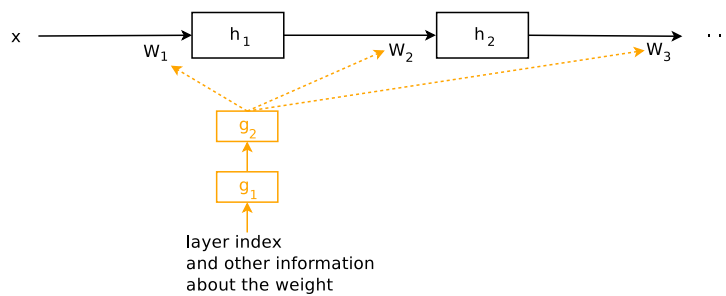

Figure 1: A hypernetwork generates the weights for a feedforward network. Black connections and parameters are associated the main network whereas orange connections and parameters are associated with the hypernetwork.

HyperNEAT (Stanley et al., 2009) is an example of hypernetworks where the inputs are a set of virtual coordinates for each weight in the main network. In this work, we will focus on a more powerful approach where the input is an embedding vector that describes the entire weights of a given layer. Our embedding vectors can be fixed parameters that are also learned during end-to-end training, allowing approximate weight-sharing within a layer and across layers of the main network. In addition, our embedding vectors can also be *generated* dynamically by our hypernetwork, allowing the weights of a recurrent network to change over timesteps and also adapt to the input sequence.

---

*Work done as a member of the Google Brain Residency program (`g.co/brainresidency`).

We perform experiments to investigate the behaviors of hypernetworks in a range of contexts and find that hypernetworks mix well with other techniques such as batch normalization and layer normalization. Our main result is that hypernetworks can generate non-shared weights for LSTM that work better than the standard version of LSTM (Hochreiter & Schmidhuber, 1997). On language modelling tasks with Character Penn Treebank, Hutter Prize Wikipedia datasets, hypernetworks for LSTM achieve near state-of-the-art results. On a handwriting generation task with IAM handwriting dataset, Hypernetworks for LSTM achieves high quantitative and qualitative results. On image classification with CIFAR-10, hypernetworks, when being used to generate weights for a deep convnet (LeCun et al., 1990), obtain respectable results compared to state-of-the-art models while having fewer learnable parameters. In addition to simple tasks, we show that Hypernetworks for LSTM offers an increase in performance for large, production-level neural machine translation models.

## 2 MOTIVATION AND RELATED WORK

Our approach is inspired by methods in evolutionary computing, where it is difficult to directly operate in large search spaces consisting of millions of weight parameters. A more efficient method is to evolve a smaller network to generate the structure of weights for a larger network, so that the search is constrained within the much smaller weight space. An instance of this approach is the work on the HyperNEAT framework (Stanley et al., 2009). In the HyperNEAT framework, Compositional Pattern-Producing Networks (CPPNs) are evolved to define the weight structure of much larger main network. Closely related to our approach is a simplified variation of HyperNEAT, where the structure is fixed and the weights are evolved through Discrete Cosine Transform (DCT) is called Compressed Weight Search (Koutnik et al., 2010). Even more closely related to our approach are Differentiable Pattern Producing Networks (DPPNs), where the structure is evolved but the weights are learned (Fernando et al., 2016), and ACDC-Networks (Moczulski et al., 2015), where linear layers are compressed with DCT and the parameters are learned.

Most reported results using these methods, however, are in small scales, perhaps because they are both slow to train and require heuristics to be efficient. The main difference between our approach and HyperNEAT is that hypernetworks in our approach are trained end-to-end with gradient descent together with the main network, and therefore are more efficient.

In addition to end-to-end learning with gradient descent, our approach strikes a good balance between Compressed Weight Search and HyperNEAT in terms of model flexibility and training simplicity. First, it can be argued that Discrete Cosine Transform used in Compressed Weight Search may be too simple and using the DCT prior may not be suitable for many problems. Second, even though HyperNEAT is more flexible, evolving both the architecture and the weights in HyperNEAT is often an overkill for most practical problems.

Even before the work on HyperNEAT and DCT, Schmidhuber (1992; 1993) has suggested the concept of fast weights in which one network can produce context-dependent weight changes for a second network. Small scale experiments were conducted to demonstrate fast weights for feed forward networks at the time, but perhaps due to the lack of modern computational tools, the recurrent network version was mentioned mainly as a thought experiment (Schmidhuber, 1993). A subsequent work demonstrated practical applications of fast weights (Gomez & Schmidhuber, 2005), where a generator network is learnt through evolution to solve an artificial control problem. The concept of a network interacting with another network is central to the work of (Jaderberg et al., 2016; Andrychowicz et al., 2016), and especially (Denil et al., 2013; Yang et al., 2015; Bertinetto et al., 2016; De Brabandere et al., 2016), where certain parameters in a convolutional network are predicted by another network. These studies however did not explore the use of this approach to recurrent networks, which is a main contribution of our work.

The focus of this work is to generate weights for practical architectures, such as convolutional networks and recurrent networks by taking layer embedding vectors as inputs. However, our hypernetworks can also be utilized to generate weights for a fully connected network by taking coordinate information as inputs similar to DPPNs. Using this setting, hypernetworks can approximately re-cover the convolutional architecture without explicitly being told to do so, a similar result obtained by "Convolution by Evolution" (Fernando et al., 2016). This result is described in Appendix A.1.

## 3 METHODS

In this paper, we view convolutional networks and recurrent networks as two ends of a spectrum. On one end, recurrent networks can be seen as imposing weight-sharing across layers, which makes them inflexible and difficult to learn due to vanishing gradient. On the other end, convolutional networks enjoy the flexibility of not having weight-sharing, at the expense of having redundant parameters when the networks are deep. Hypernetworks can be seen as a form of relaxed weight-sharing, and therefore strikes a balance between the two ends. See Appendix A.2 for conceptual diagrams of Static and Dynamic Hypernetworks.

### 3.1 STATIC HYPERNETWORK: A WEIGHT FACTORIZATION APPROACH FOR DEEP CONVOLUTIONAL NETWORKS

First we will describe how we construct a hypernetwork for the purpose of generating the weights of a feedforward convolutional network. In a typical deep convolutional network, the majority of model parameters are in the kernels of convolutional layers. Each kernel contain $N_{in} \times N_{out}$ filters and each filter has dimensions $f_{size} \times f_{size}$. Let's suppose that these parameters are stored in a matrix $K^j \in \mathbb{R}^{N_{in}f_{size} \times N_{out}f_{size}}$ for each layer $j = 1, .., D$, where $D$ is the depth of the main convolutional network. For each layer $j$, the hypernetwork receives a layer embedding $z^j \in \mathbb{R}^{N_z}$ as input and predicts $K^j$, which can be generally written as follows:

$$K^j = g(z^j), \quad \forall j = 1, ..., D \tag{1}$$

We note that this matrix $K^j$ can be broken down as $N_{in}$ slices of a smaller matrix with dimensions $f_{size} \times N_{out}f_{size}$, each slice of the kernel is denoted as $K_i^j \in \mathbb{R}^{f_{size} \times N_{out}f_{size}}$. Therefore, in our approach, the hypernetwork is a two-layer linear network. The first layer of the hypernetwork takes the input vector $z^j$ and linearly projects it into the $N_{in}$ inputs, with $N_{in}$ different matrices $W_i \in \mathbb{R}^{d \times N_z}$ and bias vectors $B_i \in \mathbb{R}^d$, where $d$ is the size of the hidden layer in the hypernetwork. For our purpose, we fix $d$ to be equal to $N_z$ although they can be different. The final layer of the hypernetwork is a linear operation which takes an input vector $a_i$ of size $d$ and linearly projects that into $K_i$ using a common tensor $W_{out} \in \mathbb{R}^{f_{size} \times N_{out}f_{size} \times d}$ and bias matrix $B_{out} \in \mathbb{R}^{f_{size} \times N_{out}f_{size}}$. The final kernel $K^j$ will be a concatenation of every $K_i^j$. Thus $g(z^j)$ can be written as follows:

$$\begin{aligned}
a_i^j &= W_i z^j + B_i, & \forall i = 1, .., N_{in}, \forall j = 1, ..., D \\
K_i^j &= \langle W_{out}, a_i^j \rangle\ ^1 + B_{out}, & \forall i = 1, .., N_{in}, \forall j = 1, ..., D \\
K^j &= \begin{pmatrix} K_1^j & K_2^j & ... & K_i^j & ... & K_{N_{in}}^j \end{pmatrix}, & \forall j = 1, ..., D
\end{aligned} \tag{2}$$

In our formulation, the learnable parameters are $W_i, B_i, W_{out}, B_{out}$ together with all $z^j$'s. During inference, the model simply takes the layer embeddings $z^j$ learned during training to reproduce the kernel weights for layer $j$ in the main convolutional network. As a side effect, the number of learnable parameters in hypernetwork will be much lower than the main convolutional network. In fact, the total number of learnable parameters in hypernetwork is $N_z \times D + d \times (N_z + 1) \times N_i + f_{size} \times N_{out} \times f_{size} \times (d+1)$ compared to the $D \times N_{in} \times f_{size} \times N_{out} \times f_{size}$ parameters for the kernels of the main convolutional network.

Our approach of constructing $g(.)$ is similar to the hierarchically semiseparable matrix approach proposed by Xia et al. (2010). Note that even though it seems redundant to have a two-layered linear hypernetwork as that is equivalent to a one-layered hypernetwork, the fact that $W_{out}$ and $B_{out}$ are shared makes our two-layered hypernetwork more compact than a one-layered hypernetwork. More concretely, a one-layered hypernetwork would have $N_z \times N_{in} \times f_{size} \times N_{out} \times f_{size}$ learnable parameters which is usually much bigger than a two-layered hypernetwork does.

---
[1] Tensor dot product between $W \in \mathbb{R}^{m \times n \times d}$ and $a \in \mathbb{R}^d$. Result $\langle W, a \rangle \in \mathbb{R}^{m \times n}$

The above formulation assumes that the network architecture consists of kernels with same dimensions. In practice, deep convolutional network architectures consists of kernels of varying dimensions. Typically, in many designs, the kernel dimensions are integer multiples of a basic size. This is indeed the case in the residual network family of architectures (He et al., 2016a) that we will be experimenting with later is an example of such a design. In our experiments, although the kernels of a residual network do not share the same dimensions, the $N_i$ and $N_{out}$ dimensions for each kernel are integer multiples of 16. To modify our approach to work with this architecture, we have our hypernetwork generate kernels for this basic size of 16, and if we require a larger kernel for a certain layer, we will concatenate multiple basic kernels together to form the larger kernel.

$$K_{32 \times 64} = \begin{pmatrix} K_1 & K_2 & K_3 & K_4 \\ K_5 & K_6 & K_7 & K_8 \end{pmatrix} \tag{3}$$

For example, if we need to generate a kernel with $N_i = 32$ and $N_{out} = 64$, we will tile eight basic kernels together. Each basic kernel is generated by a unique $z$ embedding, hence the larger kernel will be expressed with eight embeddings. Therefore, kernels that are larger in size will require a proportionally larger number of embedding vectors. For visualizations of concatenated kernels, please see Appendix A.2.1. Figure 2 shows the similarity between kernels learned by a ConvNet to classify MNIST digits and those learned by a hypernetwork generating weights for a ConvNet.

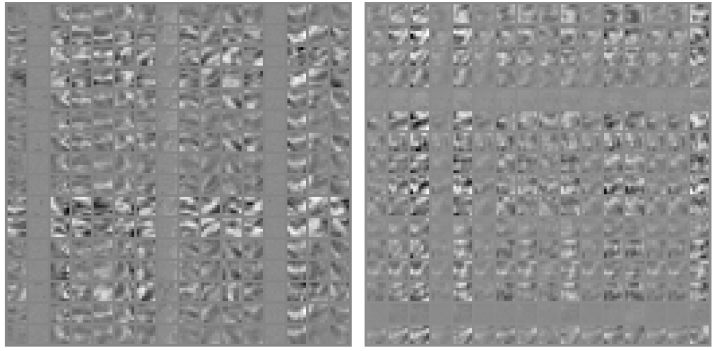

Figure 2: Kernels learned by a ConvNet to classify MNIST digits (left). Kernels learned by a hypernetwork generating weights for the ConvNet (right).

### 3.2 DYNAMIC HYPERNETWORK: ADAPTIVE WEIGHT GENERATION FOR RECURRENT NETWORKS

In the previous section, we outlined a procedure for using a hypernetwork to generate the weights for a deep convolutional network. In this section, we will use a recurrent network to dynamically generate weights for another recurrent network, such that the weights can vary across many timesteps. In this context, hypernetworks are called dynamic hypernetworks, and can be seen as a form of *relaxed* weight-sharing, a compromise between hard weight-sharing of traditional recurrent networks, and no weight-sharing of convolutional networks. This relaxed weight-sharing approach allows us to control the trade off between the number of model parameters and model expressiveness.

Our dynamic hypernetworks can be used to generate weights for RNN and LSTM. When a hypernetwork is used to generate the weights for an RNN, it is called HyperRNN. At every time step $t$, a HyperRNN takes as input the concatenated vector of input $x_t$ and the hidden states of the main RNN $h_{t-1}$, it then generates as output the vector $\hat{h}_t$. This vector is then used to generate the weights for the main RNN at the same timestep. Both the HyperRNN and the main RNN are trained jointly with backpropagation and gradient descent. In the following, we will give a more formal description of the model.

The standard formulation of a Basic RNN is given by:
$$h_t = \phi(W_h h_{t-1} + W_x x_t + b) \tag{4}$$

where $h_t$ is the hidden state, $\phi$ is a non-linear operation such as $tanh$ or $relu$, and the weight matrices and bias $W_h \in \mathbb{R}^{N_h \times N_h}, W_x \in \mathbb{R}^{N_h \times N_x}, b \in \mathbb{R}^{N_h}$ is fixed each timestep for an input sequence $X = (x_1, x_2, \ldots, x_T)$.

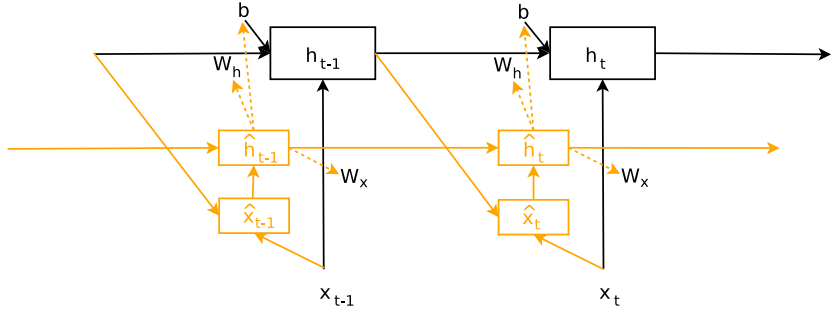

Figure 3: An overview of HyperRNNs. Black connections and parameters are associated basic RNNs. Orange connections and parameters are introduced in this work and associated with HyperRNNs. Dotted arrows are for parameter generation.

In HyperRNN, we allow $W_h$ and $W_x$ to float over time by using a smaller hypernetwork to generate these parameters of the main RNN at each step (see Figure 3). More concretely, the parameters $W_h, W_x, b$ of the main RNN are different at different time steps, so that $h_t$ can now be computed as:

$$\begin{aligned} h_t &= \phi\big(W_h(z_h)h_{t-1} + W_x(z_x) + b(z_b)\big), \text{ where} \\ W_h(z_h) &= \langle W_{hz}, z_h \rangle \\ W_x(z_x) &= \langle W_{xz}, z_x \rangle \\ b(z_b) &= W_{bz}z_b + b_0 \end{aligned} \quad (5)$$

Where $W_{hz} \in \mathbb{R}^{N_h \times N_h \times N_z}, W_{xz} \in \mathbb{R}^{N_h \times N_x \times N_z}, W_{bz} \in \mathbb{R}^{N_h \times N_z}, b_0 \in \mathbb{R}^{N_h}$ and $z_h, z_x, z_z \in \mathbb{R}^{N_z}$. We use a recurrent hypernetwork to compute $z_h, z_x$ and $z_b$ as a function of $x_t$ and $h_{t-1}$:

$$\begin{aligned} \hat{x}_t &= \begin{pmatrix} h_{t-1} \\ x_t \end{pmatrix} \\ \hat{h}_t &= \phi(W_{\hat{h}}\hat{h}_{t-1} + W_{\hat{x}}\hat{x}_t + \hat{b}) \\ z_h &= W_{\hat{h}h}\hat{h}_{t-1} + b_{\hat{h}h} \\ z_x &= W_{\hat{h}x}\hat{h}_{t-1} + b_{\hat{h}x} \\ z_b &= W_{\hat{h}b}\hat{h}_{t-1} \end{aligned} \quad (6)$$

Where $W_{\hat{h}} \in \mathbb{R}^{N_{\hat{h}} \times N_{\hat{h}}}, W_{\hat{x}} \in \mathbb{R}^{N_{\hat{h}} \times (N_h + N_z)}, b \in \mathbb{R}^{N_{\hat{h}}}$, and $W_{\hat{h}h}, W_{\hat{h}x}, W_{\hat{h}b} \in \mathbb{R}^{N_z \times N_{\hat{h}}}$ and $b_{\hat{h}h}, b_{\hat{h}x} \in \mathbb{R}^{N_z}$. This *HyperRNN Cell* has $N_{\hat{h}}$ hidden units. Typically $N_{\hat{h}}$ is much smaller than $N_h$.

As the embeddings $z_h, z_x$ and $z_b$ are of dimensions $N_z$, which is typically smaller than the hidden state size $N_{\hat{h}}$ of the HyperRNN cell, a linear network is used to project the output of the HyperRNN cell into the embeddings in Equation 6. After the embeddings are computed, they will be used to generate the full weight matrix of the main RNN.

The above is a general formulation of a *linear* dynamic hypernetwork applied to RNNs. However, we found that in practice, Equation 5 is often not practical because the memory usage becomes too large for real problems. The amount of memory required in the system described in Equation 5 will be $N_z$ times the memory of a Basic RNN, which limits the number of hidden units we can use in many practical applications.

We can modify the dynamic hypernetwork system described in Equation 5 so that it can be much more scalable and memory efficient. Our approach borrows from the static hypernetwork section and we will use an intermediate hidden vector $d(z) \in \mathbb{R}^{N_h}$ to parametrize a weight matrix, where $d(z)$ will be a linear projection of $z$. To dynamically modify a weight matrix $W$, we will allow each row of this weight matrix to be scaled linearly by an element in vector $d$. We refer $d$ as a *weight scaling vector*. Below is the modification to $W(z)$:

$$W(z) = W\big(d(z)\big) = \begin{pmatrix} d_0(z)W_0 \\ d_1(z)W_1 \\ ... \\ d_{N_h}(z)W_{N_h} \end{pmatrix} \qquad (7)$$

While we sacrifice the ability to construct an entire weight matrix from a linear combination of $N_z$ matrices of the same size, we are able to linearly scale the rows of a single matrix with $N_z$ degrees of freedom. We find this to be a good trade off, as this formulation of converting $W(z)$ into $\tilde{W}(d(z))$ decreases the amount of memory required by the dynamic hypernetwork. Rather than requiring $N_z$ times the memory of a Basic RNN, we will only be using memory in the order $N_z$ times the number of hidden units, which is an acceptable amount of extra memory usage that is often available in many applications. In addition, the row-level operation in Equation 7 can be shown to be equivalent to an element-wise multiplication operator and hence computationally much more efficient in practice. Below is the more memory efficient version of the setup of Equation 5:

$$\begin{aligned} h_t &= \phi\big(d_h(z_h) \odot W_h h_{t-1} + d_x(z_x) \odot W_x x_t + b(z_b)\big), \text{ where} \\ d_h(z_h) &= W_{hz} z_h \\ d_x(z_x) &= W_{xz} z_x \\ b(z_b) &= W_{bz} z_b + b_0 \end{aligned} \qquad (8)$$

This formulation of the HyperRNN has some similarities to Recurrent Batch Normalization (Cooijmans et al., 2016) and Layer Normalization (Ba et al., 2016). The central idea for the normalization techniques is to calculate the first two statistical moments of the inputs to the activation function, and to linearly scale the inputs to have zero mean and unit variance. An additional set of fixed parameters are learned to *unscale* the activations if required. This element-wise operation also has similarities to the Multiplicative RNN (Sutskever et al., 2011) and Multiplicative Integration RNN (Wu et al., 2016) where it was demonstrated that the multiplication-operation encouraged better gradient flow.

Since the HyperRNN cell can indirectly modify the rows of each weight matrix and also the bias of the main RNN, it is implicitly also performing a linear scaling to the inputs of the activation function. The difference here is that the linear scaling parameters can be different for each timestep and also for for each input sample. It will be interesting to compare the scaling policy that the HyperRNN cell comes up with, to the hand engineered statistical-moments based scaling approaches. In addition, we note that the existing normalization approaches can work together with the HyperRNN approach, where the HyperRNN cell will be tasked with discovering a better dynamical scaling policy to complement normalization. We will also explore this combination in our experiments.

The Long Short-Term Memory (LSTM) architecture (Hochreiter & Schmidhuber, 1997) is usually better than the Basic RNN at storing and retrieving information over longer time steps. In our experiments, we will focus on this LSTM version of the HyperRNN, called the HyperLSTM. The details of the HyperLSTM architecture is described in Appendix A.2.2, along with specific implementation details in Appendix A.2.3. We want to know whether the HyperLSTM cell can learn a weight adjustment policy that can rival statistical moments-based normalization methods, hence Layer Normalization will be one of our baseline methods. We will therefore conduct experiments on two versions of HyperLSTM, one with and one without the application of Layer Normalization.

## 4 EXPERIMENTS

In the following experiments, we will benchmark the performance of static hypernetworks on image recognition with MNIST and CIFAR-10, and the performance of dynamic hypernetworks on language modelling with Penn Treebank and Hutter Prize Wikipedia (`enwik8`) datasets and handwriting generation.

## 4.1 USING STATIC HYPERNETWORKS TO GENERATE FILTERS FOR CONVOLUTIONAL NETWORKS AND MNIST

We start by applying a hypernetwork to generate the filters for a convolutional network on MNIST. Our main convolutional network is a small two layer network and the hypernetwork is used to generate the kernel for the second layer (7x7x16x16), which contains the bulk of the trainable parameters in the system. Our weight matrix will be summarized by an embedding of size $N_z = 4$. See Appendix A.3.1 for further experimental setup details.

For this task, the hypernetwork achieved a test accuracy of 99.24%, comparable to the 99.28% for the conventional method. In this example, a kernel consisting of 12,544 weights is represented by an embedding vector of only 4 parameters, generated by a hypernetwork that has 4240 parameters. We can see the weight matrix this network produced by the hypernetwork in Figure 2. Now the question is whether we can also train a deep convolutional network, using a single hypernetwork generating a set of weights for each layer, on a dataset more challenging than MNIST.

## 4.2 STATIC HYPERNETWORKS FOR RESIDUAL NETWORK ARCHITECTURE AND CIFAR-10

The residual network architectures (He et al., 2016a; Zagoruyko & Komodakis, 2016) are popular for image recognition tasks, as they can accommodate very deep networks while maintaining effective gradient flow across layers using skip connections. The original resnet and subsequent derivatives (Zhang et al., 2016; Huang et al., 2016a) achieved state-of-the-art image recognition performance on a variety of public datasets. While residual networks can be be very deep, and in some experiments as deep as 1001 layers ((He et al., 2016b), it is important to understand whether some these layers share common properties and can be reduced effectively by introducing weight sharing. If we enforce weight-sharing across many layers of a deep feed forward network, the network may share many properties to that of a recurrent network. In this experiment, we want to explore this idea of enforcing *relaxed* weight sharing across all of the layers of a deep residual network. We will take a simple version of residual network, use a single hypernetwork to generate the weights of all of its layers for image classification task on the CIFAR-10 dataset.

| group name | output size | block type |
|---|---|---|
| conv1 | 32 × 32 | [3×3, 16] |
| conv2 | 32×32 | $\left[\begin{array}{c}3\times3, 16\times k \\ 3\times3, 16\times k\end{array}\right]\times N$ |
| conv3 | 16×16 | $\left[\begin{array}{c}3\times3, 32\times k \\ 3\times3, 32\times k\end{array}\right]\times N$ |
| conv4 | 8×8 | $\left[\begin{array}{c}3\times3, 64\times k \\ 3\times3, 64\times k\end{array}\right]\times N$ |
| avg-pool | 1 × 1 | [8 × 8] |

Table 1: Structure of Wide Residual Networks in Zagoruyko & Komodakis (2016). $N$ determines the number of residual blocks in each group. Network width is determined by factor $k$.

Our experiment will use a version of the wide residual network (Zagoruyko & Komodakis, 2016), described in Table 1, a popular and simple variant of the family of residual network architectures, and we will focus configurations ($N = 6, K = 1$) and ($N = 6, K = 2$), referred to as WRN 40-1 and WRN 40-2 respectively. In this setup, we will use a hypernetwork to generate all of the kernels in conv2, conv3, and conv4, so we will generate 36 layers of kernels in total. The WRN architecture uses a filter size of 3 for every kernel. We use the method outlined in the Methods section to deal with kernels of varying sizes, and use the an embedding size of $N_z = 64$ in our experiments. See Appendix A.3.2 for further experimental setup details.

We obtained similar classification accuracy numbers as reported in (Zagoruyko & Komodakis, 2016) with our own implementation. We also note that the weights generated by the hypernetwork are used in a batch normalization setting without modification to the original model. In principle, hypernetworks can also be applied to the newer variants of residual networks with more skip connections, such as DenseNets and ResNets of Resnets.

From the results, we see that enforcing a relaxed weight sharing constraint to the deep residual network cost us ∼ 1.25-1.5% in classification accuracy, while drastically reducing the number of parameters in the model as a trade off. One reason for this reduction in accuracy is because different layers of a deep network is trained to extract different levels of features, and require different kinds of filters to perform optimally. The hypernetwork enforces some commonality between every layer, but offers each layer 64 degrees of freedom to distinguish itself from the other layers. While the network is no longer able to learn the optimal set of filters for each layer, it will learn the best set of filters given the constraints, and the resulting number of model parameters is drastically reduced.

| Model | Test Error | Param Count |
|---|---|---|
| Network in Network (Lin et al., 2014) | 8.81% | |
| FitNet (Romero et al., 2014) | 8.39% | |
| Deeply Supervised Nets (Lee et al., 2015) | 8.22% | |
| Highway Networks (Srivastava et al., 2015) | 7.72% | |
| ELU (Clevert et al., 2015) | 6.55% | |
| Original Resnet-110 (He et al., 2016a) | 6.43% | 1.7 M |
| Stochastic Depth Resnet-110 (Huang et al., 2016b) | 5.23% | 1.7 M |
| Wide Residual Network 40-1 (Zagoruyko & Komodakis, 2016) | 6.85% | 0.6 M |
| Wide Residual Network 40-2 (Zagoruyko & Komodakis, 2016) | 5.33% | 2.2 M |
| Wide Residual Network 28-10 (Zagoruyko & Komodakis, 2016) | 4.17% | 36.5 M |
| ResNet of ResNet 58-4 (Zhang et al., 2016) | 3.77% | 13.3 M |
| DenseNet (Huang et al., 2016a) | 3.74% | 27.2 M |
| Wide Residual Network 40-1[2] | 6.73% | 0.563 M |
| Hyper Residual Network 40-1 (ours) | 8.02% | 0.097 M |
| Wide Residual Network 40-2[2] | 5.66% | 2.236 M |
| Hyper Residual Network 40-2 (ours) | 7.23% | 0.148 M |

Table 2: CIFAR-10 Classification with hypernetwork generated weights.

### 4.3 HYPERLSTM FOR CHARACTER-LEVEL PENN TREEBANK LANGUAGE MODELLING

The HyperLSTM model is evaluated on character level prediction task on the Penn Treebank corpus (Marcus et al., 1993) using the train/validation/test split outlined in (Mikolov et al., 2012). As the dataset is quite small is prone to over fitting, we apply dropout on both input and output layers with a keep probability of 0.90. Unlike previous approaches (Graves, 2013; Ognawala & Bayer, 2014) of applying weight noise during training, we instead also apply dropout to the recurrent layer (Henaff et al., 2016) with the same dropout probability.

We compare our model to the basic LSTM cell, stacked LSTM cells (Graves, 2013), and LSTM with layer normalization applied. In addition, we also experimented with applying layer normalization to HyperLSTM. Using the setup in (Graves, 2013), we use networks with 1000 units and train the network to predict the next character. In this task, the HyperLSTM cell has 128 units and a signal size of 4. As the HyperLSTM cell has more trainable parameters compared to the basic LSTM Cell, we also experimented with an LSTM Cell with 1250 units as well. For more details regarding experimental setup, please refer to Appendix A.3.3

It is interesting to note that combining Recurrent Dropout with a basic LSTM cell achieves quite formidable performance. Our implementation of Recurrent Dropout Basic LSTM cell reproduced similar results as (Semeniuta et al., 2016), where they have also experimented with different dropout settings. We also found that Layer Norm LSTM performed quite well when combined with recurrent dropout, making it both a formidable baseline and also an extension for HyperLSTM.

In addition to outperforming both the larger or deeper version of the LSTM network, HyperLSTM also achieved similar performance of Layer Norm LSTM. This suggests by dynamically adjusting the weight scaling vectors, the HyperLSTM cell has learned a policy of scaling inputs to the activation functions that is as efficient as the statistical moments-based strategy employed by Layer Norm, and that the required extra computation required is embedded inside the extra 128 units inside the HyperLSTM cell. When we combine HyperLSTM with Layer Norm, we see an additional performance gain, implying that the HyperLSTM cell learned an adjustment policy that goes beyond moments-based regularization. We also demonstrate that increasing the size of the embedding vector or stacking HyperLSTM layers together can also increase its performance.

| Model[1] | Test | Validation | Param Count |
|---|---|---|---|
| ME n-gram (Mikolov et al., 2012) | 1.37 | | |
| Batch Norm LSTM (Cooijmans et al., 2016) | 1.32 | | |
| Recurrent Dropout LSTM (Semeniuta et al., 2016) | 1.301 | 1.338 | |
| Zoneout RNN (Krueger et al., 2016) | 1.27 | | |
| HM-LSTM[3] (Chung et al., 2016) | 1.27 | | |
| LSTM, 1000 units [2] | 1.312 | 1.347 | 4.25 M |
| LSTM, 1250 units[2] | 1.306 | 1.340 | 6.57 M |
| 2-Layer LSTM, 1000 units[2] | 1.281 | 1.312 | 12.26 M |
| Layer Norm LSTM, 1000 units[2] | 1.267 | 1.300 | 4.26 M |
| HyperLSTM (ours), 1000 units | 1.265 | 1.296 | 4.91 M |
| Layer Norm HyperLSTM, 1000 units (ours) | 1.250 | 1.281 | 4.92 M |
| Layer Norm HyperLSTM, 1000 units, Large Embedding (ours) | 1.233 | 1.263 | 5.06 M |
| 2-Layer Norm HyperLSTM, 1000 units | 1.219 | 1.245 | 14.41 M |

Table 3: Bits-per-character on the Penn Treebank test set.

### 4.4 HYPERLSTM FOR HUTTER PRIZE WIKIPEDIA LANGUAGE MODELLING

We train our model on the larger and more challenging Hutter Prize Wikipedia dataset, also known as `enwik8` (Hutter, 2012) consisting of a sequence of 100M characters composed of 205 unique characters. Unlike Penn Treebank, `enwik8` contains some foreign words (Latin, Arabic, Chinese), indented XML, metadata, and internet addresses, making it a more realistic and practical dataset to test character language models. For more details regarding experimental setup, please refer to Appendix A.3.4. Examples of these mixed variety of text samples that our HyperLSTM model can generate is in Appendix A.4.

| Model[1] | `enwik8` | Param Count |
|---|---|---|
| Stacked LSTM (Graves, 2013) | 1.67 | 27.0 M |
| MRNN (Sutskever et al., 2011) | 1.60 | |
| GF-RNN (Chung et al., 2015) | 1.58 | 20.0 M |
| Grid-LSTM (Kalchbrenner et al., 2016) | 1.47 | 16.8 M |
| LSTM (Rocki, 2016b) | 1.45 | |
| MI-LSTM (Wu et al., 2016) | 1.44 | |
| Recurrent Highway Networks (Zilly et al., 2016) | 1.42 | 8.0 M |
| Recurrent Memory Array Structures (Rocki, 2016a) | 1.40 | |
| HM-LSTM[3] (Chung et al., 2016) | 1.40 | |
| Surprisal Feedback LSTM[4] (Rocki, 2016b) | 1.37 | |
| LSTM, 1800 units, no recurrent dropout[2] | 1.470 | 14.81 M |
| LSTM, 2000 units, no recurrent dropout[2] | 1.461 | 18.06 M |
| Layer Norm LSTM, 1800 units[2] | 1.402 | 14.82 M |
| HyperLSTM (ours), 1800 units | 1.391 | 18.71 M |
| Layer Norm HyperLSTM, 1800 units (ours) | 1.353 | 18.78 M |
| Layer Norm HyperLSTM, 2048 units (ours) | 1.340 | 26.54 M |

Table 4: Bits-per-character on the `enwik8` test set.

We see that HyperLSTM is once again competitive to Layer Norm LSTM, and if we combine both techniques, the Layer Norm HyperLSTM achieves respectable results. The version of HyperLSTM that uses 2048 hidden units achieve near state-of-the-art performance for this task. In addition, HyperLSTM converges quicker per training step compared to LSTM and Layer Norm LSTM. Please refer to Figure 6 for the loss graphs.

---

[1] We do not compare against methods that use dynamic evaluation.

[2] Our implementation.

[3] Based on results of version 2 at the time of writing. http://arxiv.org/abs/1609.01704v2

[4] This method uses information about test errors during inference for predicting the next characters, hence it is not directly comparable to other methods that do not use this information.

![Example text generated from HyperLSTM with weight change visualization below each character]

Figure 4: Example text generated from HyperLSTM model. We visualize how four of the main RNN's weight matrices ($W_h^i$, $W_h^g$, $W_h^f$, $W_h^o$) effectively change over time by plotting the norm of the changes below each generated character. High intensity represent large changes being made to weights of main RNN.

When we use this prediction model as a generative model to sample a text passage, we use main RNN to model a probability distribution over possible characters conditioned over the preceding characters. In the case of the HyperRNN, we allow the *model parameters* of this generative model to vary over time, so in effect the HyperRNN cell is choosing the best model at any given time to generate a probability distribution to sample from. We can demonstrate this by visualizing how the weight scaling vectors of the main RNN change during the character sampling process. In Figure 4, we examine a sample text passage generated by HyperLSTM after training on `enwik8` along with the weight differences below the text. We see that in regions of low intensity, where the weights of the main RNN are relatively static, the types of phrases generated seem more deterministic. For example, the weights do not change much during the words `Europeans`, `possessions` and `reservation`. The regions of high intensity is when the HyperRNN cell is making relatively large changes to the weights of the main RNN. These tend to happen in the areas between words, or sometimes during brackets.

One might also wonder whether the HyperLSTM cell (without Layer Norm), via dynamically tuning the weight scaling vectors, has developed a policy that is similar to the statistics-based approach used by Layer Norm, given that both methods have similar performance. One way to see this effect is to look at the histogram of the hidden states in the network. In Figure 5, we examine the histograms of $\phi(c_t)$, the hidden state of the LSTM before applying the output gate.

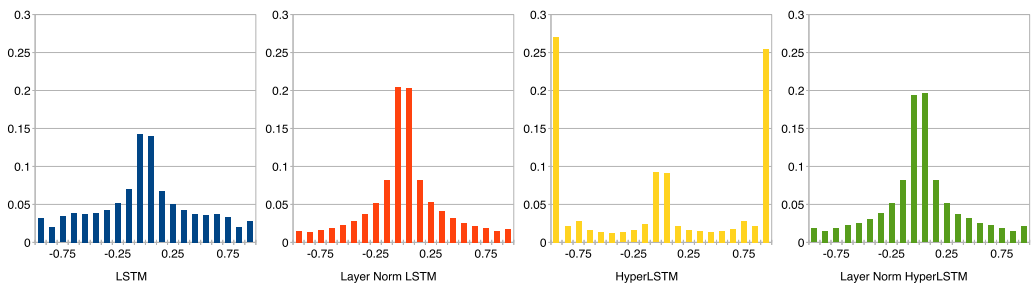

Figure 5: Normalized Histogram plots of $\phi(c_t)$ for different models during sampling.

We see that the normalization process employed by Layer Norm reduces the saturation effects compared to the vanilla LSTM. However, for the case of the HyperLSTM, we notice that most of the time the cell is saturated. The HyperLSTM cell's dynamic weight adjustment policy appears to be doing something very different compared to statistical normalization, although the policy it came up with ended up providing similar performance as Layer Norm. It is interesting to see that when we combine both methods, the HyperLSTM cell will need to determine an adjustment policy *in spite of* the normalization forced upon it by Layer Norm. An interesting question is whether there are problems where statistical normalization may actually be a setback to the policy developed by the HyperLSTM, and the best strategy is to ignore it.

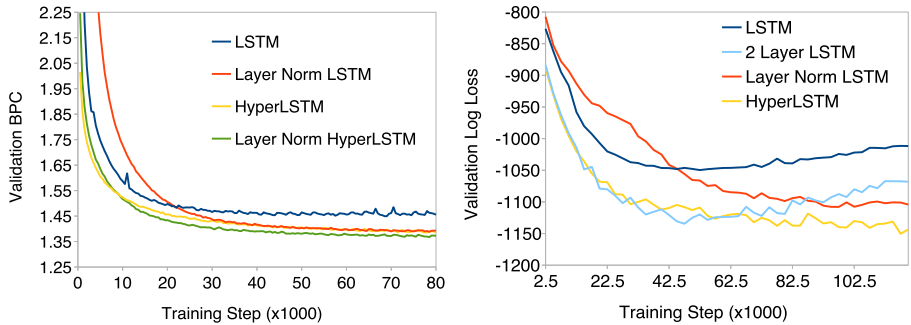

Figure 6: Loss Graph for `enwik8` (left). Loss Graph for Handwriting Generation (right)

4.5 HYPERLSTM FOR HANDWRITING SEQUENCE GENERATION

In addition to modelling discrete sequential data, we want to see how the model performs when modelling sequences of real valued data. We will train our model on the IAM online handwriting database (Liwicki & Bunke, 2005) and have our model predict pen strokes as per Section 4.2 of (Graves, 2013). The dataset has contains 12179 handwritten lines from 221 writers, digitally recorded from a tablet. We will model the (x, y) coordinate of the pen location at each recorded time step, along with a binary indicator of pen-up/pen-down. The average sequence length is around 700 steps and the longest around 1900 steps, making the training task particularly challenging as the network needs to retain information about both the stroke history and also the handwriting style in order to predict plausible future handwriting strokes. For experimental setup details, please refer to Appendix A.3.5.

| Model | Log-Loss | Param Count |
|---|---|---|
| LSTM, 900 units (Graves, 2013) | -1,026 | |
| 3-Layer LSTM, 400 units (Graves, 2013) | -1,041 | |
| 3-Layer LSTM, 400 units, adaptive weight noise (Graves, 2013) | -1,058 | |
| LSTM, 900 units, no dropout, no data augmentation.[1] | -1,026 | 3.36 M |
| 3-Layer LSTM, 400 units, no dropout, no data augmentation.[1] | -1,039 | 3.26 M |
| LSTM, 900 units[2] | -1,055 | 3.36 M |
| LSTM, 1000 units[2] | -1,048 | 4.14 M |
| 3-Layer LSTM, 400 units[2] | -1,068 | 3.26 M |
| 2-Layer LSTM, 650 units[2] | -1,135 | 5.16 M |
| Layer Norm LSTM, 900 units[2] | -1,096 | 3.37 M |
| Layer Norm LSTM, 1000 units[2] | -1,106 | 4.14 M |
| Layer Norm HyperLSTM, 900 units (ours) | -1,067 | 3.95 M |
| HyperLSTM (ours), 900 units | -1,162 | 3.94 M |

Table 5: Log-Loss of IAM Online DB validation set.

In this task, we note that data augmentation and applying recurrent dropout improved the performance of all models, compared to the original setup by (Graves, 2013). In addition, for the LSTM model, increasing unit count per layer may not help the performance compared to increasing the layer depth. We notice that a 3-layer 400 unit LSTM outperforms a 1-layer 900 unit one, and we found that a 2-layer 650 unit LSTM outperforming most configurations. While layer norm helps with the performance, we found that in this task, layer norm does not combine well with HyperLSTM, and in this task the 900 unit HyperLSTM without layer norm achieved the best performance.

Unlike the language modelling task, perhaps statistical normalization is far from the optimal approach for a weight adjustment policy. The policy learned by the HyperLSTM cell not only per-formed well against the baseline, its convergence rate is also as fast as the 2-layer LSTM model. Please refer to Figure 6 for the loss graphs.

---

[1] Our implementation, to replicate setup of (Graves, 2013).
[2] Our implementation, with data augmentation, dropout and recurrent dropout.

In Appendix A.5, we display three sets of handwriting samples generated from LSTM, Layer Norm LSTM, and HyperLSTM, corresponding to log-loss scores of -1055, -1096, and -1162 nats respectively in Table 5. Qualitative assessments of handwriting quality is always subjective, and depends an individual's taste in calligraphy. From looking at the examples produced by the three models, our opinion is that the samples produced by LSTM is noisier than the other two models. We also find HyperLSTM's samples to be a bit more coherent than the samples produced by Layer Norm LSTM. We leave to the reader to judge which model produces handwriting samples of higher quality.

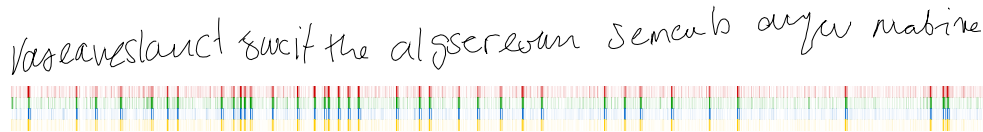

Figure 7: Handwriting sample generated from HyperLSTM model. We visualize how four of the main RNN's weight matrices ($W_h^i$ $W_h^g$, $W_h^f$, $W_h^o$) effectively change over time, by plotting norm of changes made to them over time.

Similar to the earlier character generation experiment, we show a generated handwriting sample from the HyperLSTM model in Figure 7, along with a plot of how the weight scaling vectors of the main RNN is changing over time below the sample. For a more detailed interactive demonstration of handwriting generation using HyperLSTM, visit http://blog.otoro.net/2016/09/28/hyper-networks/.

We see that the regions of high intensity seem to be concentrated at many discrete instances, rather than slowly varying over time. This implies that the weights experience regime changes rather than gradual slow adjustments. We can see that many of these weight changes occur between the written words, and sometimes between written characters. While the LSTM model alone already does a formidable job of generating time-varying parameters of a Mixture Gaussian distribution used to generate realistic handwriting samples, the ability to go one level deeper, and to dynamically generate the generative model is one of the key advantages of HyperRNN over a normal RNN.

4.6 HyperLSTM for Neural Machine Translation

We experiment with the Neural Machine Translation task using the same experimental setup outlined in (Wu et al., 2016). Our model is the same wordpiece model architecture with a vocabulary size of 32k, but we replace the LSTM cells with HyperLSTM cells. We benchmark the modified model on WMT'14 En→Fr using the same test/validation set split described in the GNMT paper (Wu et al., 2016). Please refer to Appendix A.3.6 for experimental setup details.

| Model | Test BLEU | Log Perplexity |
|---|---|---|
| Deep-Att + PosUnk (Zhou et al., 2016) | 39.2 | |
| GNMT WPM-32K, LSTM (Wu et al., 2016) | 38.95 | 1.027 |
| GNMT WPM-32K, ensemble of 8 LSTMs (Wu et al., 2016) | 40.35 | |
| GNMT WPM-32K, HyperLSTM (ours) | 40.03 | 0.993 |

Table 6: Single model results on WMT En→Fr (newstest2014)

The HyperLSTM cell improves the performance of the existing GNMT model, achieving state-of-the-art single model results for this dataset. In addition, we demonstrate the applicability of hypernetworks to large-scale models used in production systems. Please see Appendix A.6 for actual translation samples generated from both models for a qualitative comparison.

## 5 CONCLUSION

In this paper, we presented a method to use a hypernetwork to generate weights for another neural network. Our hypernetworks are trained end-to-end with backpropagation and therefore are efficient and scalable. We focused on two use cases of hypernetworks: static hypernetworks to generate weights for a convolutional network, dynamic hypernetworks to generate weights for recurrent networks. We found that the method works well while using fewer parameters. On image recognition, language modelling and handwriting generation, hypernetworks are competitive to or sometimes better than state-of-the-art models.

### ACKNOWLEDGMENTS

We thank Jeff Dean, Geoffrey Hinton, Mike Schuster and the Google Brain team for their help with the project.

# A APPENDIX

## A.1 HYPERNETWORKS TO LEARN FILTERS FOR A FULLY CONNECTED NETWORKS

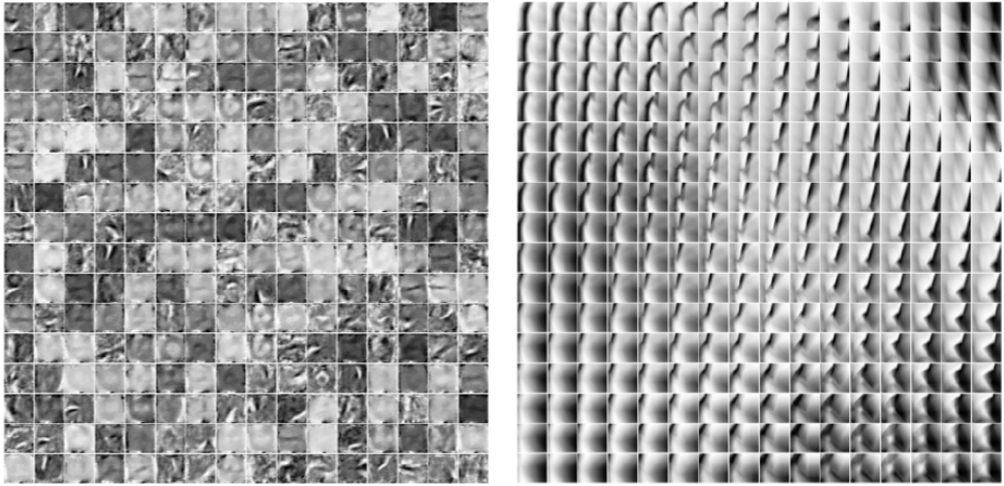

Figure 8: Filters learned to classify MNIST digits in a fully connected network (left). Filters learned by a hypernetwork (right).

We ran an experiment where the hypernetwork receives the $x, y$ locations of both the input pixel and the weight, and predicts the value of the hidden weight matrix in a fully connected network that learns to classify MNIST digits. In this experiment, the fully connected network (784-256-10) has one hidden layer of $16 \times 16$ units, where the hypernetwork is a pre-defined small feedforward network. The weights of the hidden layer has $784 \times 256 = 200704$ parameters, while the hypernetwork is a $801$ parameter four layer feed forward relu network that would generate the $786 \times 256$ weight matrix. The result of this experiment is shown in Figure 8. We want to emphasize that even though the network can learn convolutional-like filters during end-to-end training, its performance is rather poor: the best accuracy is 93.5%, compared to 98.5% for the conventional fully connected network.

We find that the virtual coordinates-based approach to hypernetworks that is used by HyperNEAT and DPPN has its limitations in many practical tasks, such as image recognition and language modelling, and therefore developed our embedding vector approach in this work.

## A.2 CONCEPTUAL DIAGRAMS OF STATIC AND DYNAMIC HYPERNETWORKS

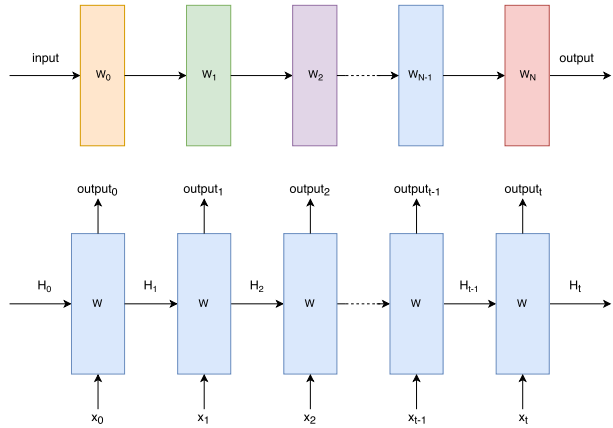

Figure 9: Feedforward Network (top) and Recurrent Network (bottom)

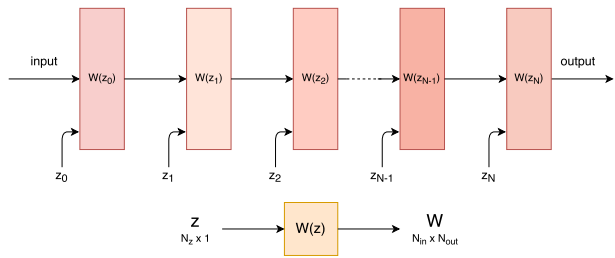

Figure 10: Static Hypernetwork generating weights for Feedforward Network

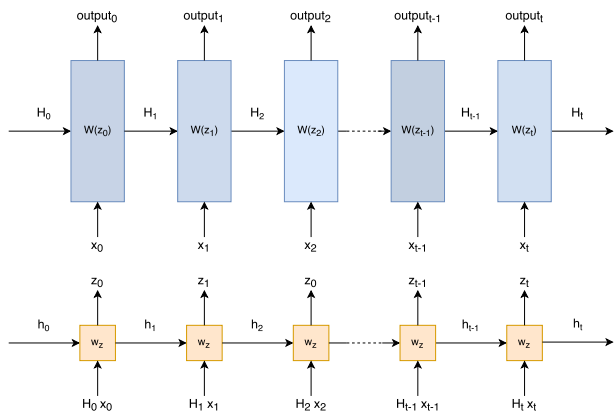

Figure 11: Dynamic Hypernetwork generating weights for Recurrent Network

### A.2.1 FILTER VISUALIZATIONS FOR RESIDUAL NETWORKS

In Figures 12 and 13 are example visualizations for various kernels in a deep residual network. Note that the 32x32x3x3 kernel generated by the hypernetwork was constructed by concatenating 4 basic kernels together.

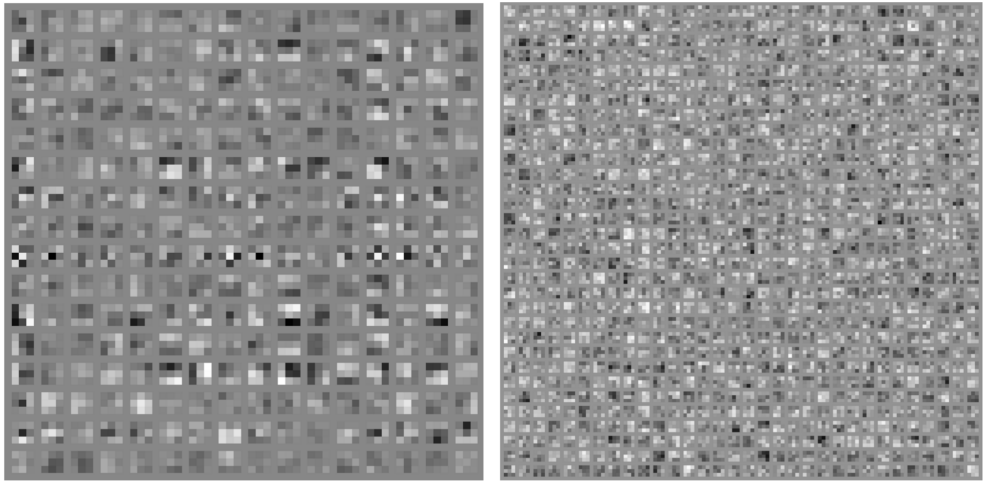

Figure 12: Normal CIFAR-10 16x16x3x3 kernel (left). Normal CIFAR-10 32x32x3x3 kernel (right).

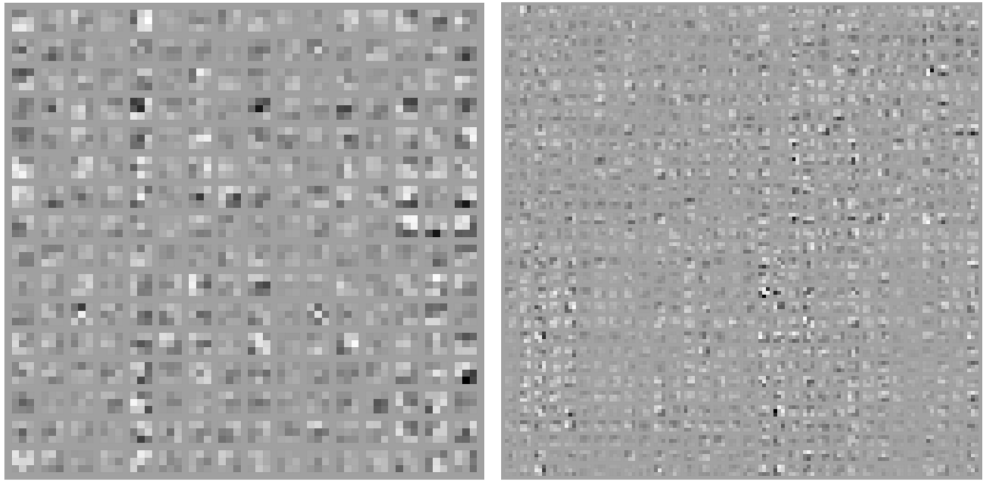

Figure 13: Generated 16x16x3x3 kernel (left). Generated 32x32x3x3 kernel (right).

### A.2.2 HYPERLSTM

In this section we will discuss extension of HyperRNN to LSTM. Our focus will be on the basic version of the LSTM architecture Hochreiter & Schmidhuber (1997), given by:

$$\begin{aligned}
i_t &= W_h^i h_{t-1} + W_x^i x_t + b^i \\
g_t &= W_h^g h_{t-1} + W_x^g x_t + b^g \\
f_t &= W_h^f h_{t-1} + W_x^f x_t + b^f \\
o_t &= W_h^o h_{t-1} + W_x^o x_t + b^o \\
c_t &= \sigma(f_t) \odot c_{t-1} + \sigma(i_t) \odot \phi(g_t) \\
h_t &= \sigma(o_t) \odot \phi(c_t)
\end{aligned} \quad (9)$$

where $W_h^y \in \mathbb{R}^{N_h \times N_h}, W_x^y \in \mathbb{R}^{N_h \times N_x}, b^y \in \mathbb{R}^{N_h}$, $\sigma$ is the *sigmoid* operator, $\phi$ is the *tanh* operator. For brevity, $y$ is one of $\{i, g, f, o\}$.[1]

Similar to the previous section, we will make the weights and biases a function of an embedding, and the embedding for each $\{i, g, f, o\}$ will be generated from a smaller HyperLSTM cell. As discussed earlier, we will also experiment with adding the option to use a Layer Normalization layer in the HyperLSTM. The HyperLSTM Cell is given by:

$$\begin{aligned}
\hat{x}_t &= \begin{pmatrix} h_{t-1} \\ x_t \end{pmatrix} \\
\hat{i}_t &= LN(W_{\hat{h}}^{\hat{i}} \hat{h}_{t-1} + W_{\hat{x}}^{\hat{i}} \hat{x}_t + \hat{b}^{\hat{i}}) \\
\hat{g}_t &= LN(W_{\hat{h}}^{\hat{g}} \hat{h}_{t-1} + W_{\hat{x}}^{\hat{g}} \hat{x}_t + \hat{b}^{\hat{g}}) \\
\hat{f}_t &= LN(W_{\hat{h}}^{\hat{f}} \hat{h}_{t-1} + W_{\hat{x}}^{\hat{f}} \hat{x}_t + \hat{b}^{\hat{f}}) \\
\hat{o}_t &= LN(W_{\hat{h}}^{\hat{o}} \hat{h}_{t-1} + W_{\hat{x}}^{\hat{o}} \hat{x}_t + \hat{b}^{\hat{o}}) \\
\hat{c}_t &= \sigma(\hat{f}_t) \odot \hat{c}_{t-1} + \sigma(\hat{i}_t) \odot \phi(\hat{g}_t) \\
\hat{h}_t &= \sigma(\hat{o}_t) \odot \phi(LN(\hat{c}_t))
\end{aligned} \quad (10)$$

The weight matrices for each of the four $\{i, g, f, o\}$ gates will be a function of a set of embeddings $z_x, z_h$, and $z_b$ unique to each gates, just like the HyperRNN. These embeddings are linear projections of the hidden states of the HyperLSTM Cell. For brevity, $y$ is one of $\{i, g, f, o\}$ to avoid writing four sets of identical equations:

$$\begin{aligned}
z_h^y &= W_{\hat{h}h}^y \hat{h}_{t-1} + b_{\hat{h}h}^y \\
z_x^y &= W_{\hat{h}x}^y \hat{h}_{t-1} + b_{\hat{h}x}^y \\
z_b^y &= W_{\hat{h}b}^y \hat{h}_{t-1}
\end{aligned} \quad (11)$$

As in the memory efficient version of the HyperRNN, we will focus on the efficient version of the HyperLSTM, where we use weight scaling vectors $d$ to modify the rows of the weight matrices:

$$\begin{aligned}
y_t &= LN\big(d_h^y(z_h) \odot W_h^y h_{t-1} + d_x^y(z_x) \odot W_x^y x_t + b^y(z_b^y)\big), \text{ where} \\
d_h^y(z_h) &= W_{hz}^y z_h \\
d_x^y(z_x) &= W_{xz}^y z_x \\
b^y(z_b^y) &= W_{bz}^y z_b^y + b_0^y
\end{aligned} \quad (12)$$

In our implementation, the cell and hidden state update equations for the main LSTM will incorporate a single dropout (Hinton et al., 2012) gate, as developed in Recurrent Dropout without Memory Loss (Semeniuta et al., 2016), as we found this to help regularize the entire model during training:

$$\begin{aligned}
c_t &= \sigma(f_t) \odot c_{t-1} + \sigma(i_t) \odot DropOut(\phi(g_t)) \\
h_t &= \sigma(o_t) \odot \phi(LN(c_t))
\end{aligned} \quad (13)$$

---
[1] In practice, all eight weight matrices are concatenated into one large matrix for computational efficiency.

This dropout operation is generally only applied inside the main LSTM, not in the smaller HyperLSTM cell. For larger size systems we can apply dropout to both networks.

### A.2.3 IMPLEMENTATION DETAILS AND WEIGHT INITIALIZATION FOR HYPERLSTM

This section may be useful to readers who may want to implement their own version of the HyperLSTM Cell, as we will discuss initialization of the parameters for Equations 10 to 13. We recommend implementing the HyperLSTM within the same interface as a normal recurrent network cell so that using the HyperLSTM will not be any different than using a normal RNN. These initialization parameters have been found to work well with our experiments, but they may be far from optimal depending on the task at hand. A reference implementation developed using the TensorFlow (Abadi et al., 2016) framework can be found at http://blog.otoro.net/2016/09/28/hyper-networks/.

The HyperLSTM Cell will be located inside the HyperLSTM, as described in Equation 10. It is a normal LSTM cell with Layer Normalization. The inputs to the HyperLSTM Cell will be the concatenation of the input signal and the hidden units of the main LSTM cell. The biases in Equation 10 are initialized to zero and Orthogonal Initialization (Henaff et al., 2016) is performed for all weights.

The embedding vectors are produced by the HyperLSTM Cell at each timestep by linear projection described in Equation 11. The weights for the first two equations are initialized to be zero, and the biases are initialized to one. The weights for the third equation are initialized to be a small normal random variable with standard deviation of 0.01.

The weight scaling vectors that modify the weight matrices are generated from these embedding vectors, as per Equation 12. Orthogonal initialization is applied to the $W_h$ and $W_x$, while $b_0$ is initialized to zero. $W_{bz}$ is also initialized to zero. For the weight scaling vectors, we used a method described in Recurrent Batch Normalization (Cooijmans et al., 2016) where the scaling vectors are initialized to 0.1 rather than 1.0 and this has shown to help gradient flow. Therefore, for weight matrices $W_{hz}$ and $W_{xz}$, we initialize to a constant value of $0.1/N_z$ to maintain this property.

The only place we use dropout is in the single location in Equation 13, developed in Recurrent Dropout without Memory Loss (Semeniuta et al., 2016). We can use this dropout gate like any other normal dropout gate in a feed-forward network.

## A.3 EXPERIMENT SETUP DETAILS AND HYPER PARAMETERS

### A.3.1 USING STATIC HYPERNETWORKS TO GENERATE FILTERS FOR CONVOLUTIONAL NETWORKS AND MNIST

We train the network with a 55000 / 5000 / 10000 split for the training, validation and test sets and use the 5000 validation samples for early stopping, and train the network using Adam (Kingma & Ba, 2015) with a learning rate of 0.001 on mini-batches of size 1000. To decrease over fitting, we pad MNIST training images to 30x30 pixels and random crop to 28x28.[1]

| Model | Test Error | Params of 2$^{nd}$ Kernel |
|---|---|---|
| Normal Convnet | 0.72% | 12,544 |
| Hyper Convnet | 0.76% | 4,244 |

Table 7: MNIST Classification with hypernetwork generated weights.

### A.3.2 STATIC HYPERNETWORKS FOR RESIDUAL NETWORK ARCHITECTURE AND CIFAR-10

We train both the normal residual network and the hypernetwork version using a 45000 / 5000 / 10000 split for training, validation, and test set. The 5000 validation samples are randomly chosen and isolated from the original 50000 training samples. We train the entire setup with a mini-batch size of 128 using Nesterov Momentum SGD for the normal version and Adam for the hypernetwork version, both with a learning rate schedule. We apply L2 regularization on the kernel weights, and also on the hypernetwork-generated kernel weights of 0.0005%. To decrease over fitting, we apply light data augmentation pad training images to 36x36 pixels and random crop to 32x32, and perform random horizontal flips.

---
[1] An IPython notebook demonstrating the MNIST Hypernetwork experiment is available at this website: http://blog.otoro.net/2016/09/28/hyper-networks/.

Table 8: Learning Rate Schedule for Nesterov Momentum SGD

| <step   | learning rate |
|---------|---------------|
| 28,000  | 0.10000       |
| 56,000  | 0.02000       |
| 84,000  | 0.00400       |
| 112,000 | 0.00080       |
| 140,000 | 0.00016       |

Table 9: Learning Rate Schedule for Hyper Network / Adam

| <step   | learning rate |
|---------|---------------|
| 168,000 | 0.00200       |
| 336,000 | 0.00100       |
| 504,000 | 0.00020       |
| 672,000 | 0.00005       |

### A.3.3 CHARACTER-LEVEL PENN TREEBANK

The hyper-parameters of all the experiments were selected through non-extensive grid search on the validation set. Whenever possible, we used reported learning rates and batch sizes in the literature that had been used for similar experiments performed in the past.

For Character-level Penn Treebank, we use mini-batches of size 128, to train on sequences of length 100. We trained the model using Adam (Kingma & Ba, 2015) with a learning rate of 0.001 and gradient clipping of 1.0. During evaluation, we generate the entire sequence, and do not use information about previous test errors for prediction, e.g., dynamic evaluation (Graves, 2013; Rocki, 2016b). As mentioned earlier, we apply dropout to the input and output layers, and also apply recurrent dropout with a keep probability of 90%. For baseline models, Orthogonal Initialization (Henaff et al., 2016) is performed for all weights.

We also experimented with a version of the model using a larger embedding size of 16, and also with a lower dropout keep probability of 85%, and reported results with this "Large Embedding" model in Table 3. Lastly, we stacked two layers of this "Large Embedding" model together to measure the benefits of a multi-layer version of HyperLSTM, with a dropout keep probability of 80%.

### A.3.4 HUTTER PRIZE WIKIPEDIA

As enwik8 is a bigger dataset compared to Penn Treebank, we will use 1800 units for our networks. In addition, we perform training on sequences of length 250. Our normal HyperLSTM Cell consists of 256 units, and we use an embedding size of 64.

Our setup is similar in the previous experiment, using the same mini-batch size, learning rate, weight initialization, gradient clipping parameters and optimizer. We do not use dropout for the input and output layers, but still apply recurrent dropout with a keep probability of 90%. For baseline models, Orthogonal Initialization (Henaff et al., 2016) is performed for all weights.

As in (Chung et al., 2015), we train on the first 90M characters of the dataset, use the next 5M as a validation set for early stopping, and the last 5M characters as the test set.

In this experiment, we also experimented with a slightly larger version of HyperLSTM with 2048 hidden units. This version of of the model uses 2048 hidden units for the main network, inline with similar models for this experiment in other works. In addition, its HyperLSTM Cell consists of 512 units with an embedding size of 64. Given the larger number of nodes in both the main LSTM and HyperLSTM cell, recurrent dropout is also applied to the HyperLSTM Cell of this model, where we use a lower dropout keep probability of 85%, and train on an increased sequence length of 300.

### A.3.5 HANDWRITING SEQUENCE GENERATION

We will use the same model architecture described in (Graves, 2013) and use a Mixture Density Network layer (Bishop, 1994) to generate a mixture of bi-variate Gaussian distributions to model at each time step to model the pen location. We normalize the data and use the same train/validation split as per (Graves, 2013) in this experiment. We remove samples less than length 300 as we found these samples contain a lot of recording errors and noise. After the pre-processing, as the dataset is small, we introduce data augmentation of chosen uniformly from +/- 10% and apply a this random scaling a the samples used for training.

One concern we want to address is the lack of a test set in the data split methodology devised in (Graves, 2013). In this task, qualitative assessment of generated handwriting samples is arguably just as important as the quantitative log likelihood score of the results. Due to the small size of the dataset, we want to use as large as possible the portion of the dataset to train our models in order to generate better quality handwriting samples so we can also judge our models qualitatively in addition to just examining the log-loss numbers, so for this task we will use the same training / validation split as (Graves, 2013), with a caveat that we may be somewhat over fitting to the validation set in the quantitative results. In future works, we will explore using larger datasets to conduct a more rigorous quantitative analysis.

For model training, will apply recurrent dropout and also dropout to the output layer with a keep probability of 0.95. The model is trained on mini-batches of size 32 containing sequences of variable length. We trained the model using Adam (Kingma & Ba, 2015) with a learning rate of 0.0001 and gradient clipping of 5.0. Our HyperLSTM Cell consists of 128 units and a signal size of 4. For baseline models, Orthogonal Initialization (Henaff et al., 2016) is performed for all weights.

### A.3.6 NEURAL MACHINE TRANSLATION

Our experimental procedure follows the procedure outlined in Sections 8.1 to 8.4 of the GNMT paper (Wu et al., 2016). We only performed experiments with a single model and did not conduct experiments with Reinforcement Learning or Model Ensembles as described in Sections 8.5 and 8.6 of the GNMT paper.

The GNMT paper outlines several methods for the training procedure, and investigated several approaches including combining Adam and SGD optimization methods, in addition to weight quantization schemes. In our experiment, we used only the Adam (Kingma & Ba, 2015) optimizer with the same hyperparameters described in the GNMT paper. We did not employ any quantization schemes.

We replaced LSTM cells in the GNMT WPM-32K architecture, with LayerNorm HyperLSTM cells with the same number of hidden units. In this experiment, our HyperLSTM Cell consists of 128 units with an embedding size of 32.

A.4 EXAMPLES OF GENERATED WIKIPEDIA TEXT

```
The eastern half of Russia varies from Modern to Central Europe. Due to
    similar lighting and the extent of the combination of long
    tributaries to the [[Gulf of Boston]], it is more of a private
    warehouse than the [[Austro-Hungarian Orthodox Christian and Soviet
    Union]].

==Demographic data base==

[[Image:Auschwitz controversial map.png|frame|The ''Austrian Spelling'']]
[[Image:Czech Middle East SSR chief state 103.JPG|thumb|Serbian Russia
    movement]] [[1593]]&ndash;[[1719]], and set up a law of [[
    parliamentary sovereignty]] and unity in Eastern churches.

In medieval Roman Catholicism Tuba and Spanish controlled it until the
    reign of Burgundian kings and resulted in many changes in
    multiculturalism, though the [[Crusades]], usually started following
    the [[Treaty of Portugal]], shored the title of three major powers,
    only a strong part.

[[French Marines]] (prompting a huge change in [[President of the Council
     of the Empire]], only after about [[1793]], the Protestant church,
    fled to the perspective of his heroic declaration of government and,
    in the next fifty years, [[Christianity|Christian]] and [[Jutland]].
    Books combined into a well-published work by a single R. (Sch. M.
    ellipse poem) tradition in St Peter also included 7:1, he dwell upon
    the apostle, scripture and the latter of Luke; totally unknown, a
    distinct class of religious congregations that describes in number of
     [[remor]]an traditions such as the [[Germanic tribes]] (Fridericus
    or Lichteusen and the Wales). Be introduced back to the [[
    14th century]], as related in the [[New Testament]] and in its elegant [[
    Anglo-Saxon Chronicle]], although they branch off the characteristic
    traditions which Saint [[Philip of Macedon]] asserted.

Ae also in his native countries.

In [[1692]], Seymour was barged at poverty of young English children,
    which cost almost the preparation of the marriage to him.

Burke's work was a good step for his writing, which was stopped by clergy
     in the Pacific, where he had both refused and received a position of
     successor to the throne. Like the other councillors in his will, the
     elder Reinhold was not in the Duke, and he was virtually non-father
    of Edward I, in order to recognize [[Henry II of England|Queen Enrie
    ]] of Parliament.

The Melchizedek Minister Qut]] signed the [[Soviet Union]], and forced
    Hoover to provide [[Hoover (disambiguation)|hoover]]s in [[1844]],
    [[1841]].

His work on social linguistic relations is divided to the several times
    of polity for educatinnisley is 760 Li Italians.  After Zaiti's death
    , and he was captured August 3, he witnessed a choice better by
    public, character, repetitious, punt, and future.
```

Figure 14: `enwik8` sample generated from 2048-unit Layer Norm HyperLSTM

```
== Quatitis==
:''Main article: [[sexagesimal]]''

Sexual intimacy was traditionally performed by a male race of the [[
    mitochondria]] of living things. The next geneme is used by ''
    Clitoron'' into short forms of [[sexual reproduction]]. When a
    maternal suffeach-Lashe]] to the myriad of a "master's character
    ". He recognizes the associated reflection of [[force call|
    carriers]], the [[Battle of Pois except fragile house and by
    historians who have at first incorporated his father.

==Geography==
The island and county top of Guernsey consistently has about a third of
    its land, centred on the coast subtained by mountain peels with
    mountains, squares, and lakes that cease to be links with the size
    and depth of sea level and weave in so close to lowlands.
    Strategically to the border of the country also at the southeast
    corner of the province of Denmark do not apply, but sometimes west of
     dense climates of coastal Austria and west Canada, the Flemish area
    of the continent actually inhabits [[tropical geographical transition
    ]] and transitions from [[soil]] to [[snow]] residents.]]

==Definition==
The symbols are ''quotational'' and '''distinct''' or advanced. {{ref|
    no_1}} Older readings are used for [[phrase]]s, especially, [[ancient
     Greek]], and [[Latin]] in their development process. Several
    varieties of permanent systems typically refer to [[primordial
    pleasure]] (for example, [[Pleistocene]], [[Classical antenni|Ctrum
    ]]), but its claim is that it holds the size of the coci, but is
    historically important both for import: brewing and commercial use.

A majority of cuisine specifically refers to this period, where the
    southern countries developed in the 19th century. Scotland had a
    cultural identity of or now a key church who worked between the 8th
    and 60th through 6 (so that there are small single authors of
    detailed recommendations for them and at first) rather than appearing
    , [[Adoptionism|adoptionists]] often started inscribed with the words
     distinct from two types.  On the group definition the adjective ''
    fighting'' is until Crown Violence Association]], in which the higher
     education [[motto]] (despite the resulting attack on [[medical
    treatment]]) peaked on [[15 December]], [[2005]]. At 30 percent, up
    to 50% of the electric music from the period was created by Voltaire,
     but Newton promoted the history of his life.

Publications in the Greek movie ''[[The Great Theory of Bertrand Russell
    ]]'', also kept an important part into the inclusion of ''[[The Beast
     for the Passage of Study]]'', began in [[1869]], opposite the
    existence of racial matters.  Many of Mary's religious faiths (
    including the [[Mary Sue Literature]] in the United States)
    incorporated much of Christianity within Hispanic [[Sacred text]]s.

But controversial belief must be traced back to the 1950s stated that
    their anticolonial forces required the challenge of even lingering
    wars tossing nomon before leaves the bomb in paint on the South
    Island, known as [[Quay]], facing [[Britain]], though he still holds
    to his ancestors a strong ancestor of Orthodoxy. Others explain that
    the process of reverence occurred from [[Common Hermitage]], when the
     [[Crusade|Speakers]] laid his lifespan in [[Islam]] into the north
    of Israel. At the end of the [[14th century BCE]], the citadel of [[
    Israel]] set Eisenace itself in the [[Abyssinia]]n islands, which was
     Faroe's Dominican Republic claimed by the King.
```

Figure 15: enwik8 sample generated from 2048-unit Layer Norm HyperLSTM

A.5 EXAMPLES OF RANDOMLY CHOSEN GENERATED HANDWRITING SAMPLES

Figure 16: Handwriting samples generated from LSTM

Figure 17: Handwriting samples generated from Layer Norm LSTM

*[handwritten text, illegible]*

Figure 18: Handwriting samples generated from HyperLSTM

## A.6 EXAMPLES OF RANDOMLY CHOSEN MACHINE TRANSLATION SAMPLES

We randomly selected translation samples generated from both LSTM baseline and HyperLSTM models from the WMT'14 En→Fr Test Set. Given an English phrase, we can compare between the correct French translation, the LSTM translation, and the HyperLSTM translation.

> English Input
> ```
> I was expecting to see gnashing of teeth and a fight breaking
> out at the gate .
> ```
> French (Ground Truth)
> ```
> Je m' attendais à voir des grincements de dents et une
> bagarre éclater à la porte .
> ```
> LSTM Translation
> ```
> Je m' attendais à voir des larmes de dents et un combat à la
> porte .
> ```
> HyperLSTM Translation
> ```
> Je m' attendais à voir des dents grincer des dents et une
> bataille éclater à la porte .
> ```

> English Input
> ```
> Prosecuting , Anne Whyte said : " If anyone should know not
> to the break the law , it is a criminal solicitor . "
> ```
> French (Ground Truth)
> ```
> Le procureur Anne Whyte a déclaré : « Si quelqu' un doit
> savoir qu' il ne faut pas violer la loi , c' est bien un
> avocat pénaliste . »
> ```
> LSTM Translation
> ```
> Prosecuting , Anne Whyte a dit : « Si quelqu' un doit savoir
> qu' il ne faut pas enfreindre la loi , c' est un solicitor
> criminel .
> ```
> HyperLSTM Translation
> ```
> En poursuivant , Anne Whyte a dit : « Si quelqu' un doit
> savoir ne pas enfreindre la loi , c' est un avocat criminel .
> ```

> English Input
> ```
> According to her , the CSRS was invited to a mediation and she
> asked for an additional period for consideration .
> ```
> French (Ground Truth)
> ```
> Selon elle , la CSRS a été invitée à une médiation et elle a
> demandé un délai supplémentaire pour y réfléchir .
> ```
> LSTM Translation
> ```
> Selon elle , le SCRS a été invité à une médiation et elle a demandé
> un délai supplémentaire .
> ```
> HyperLSTM Translation
> ```
> Selon elle , le SCRS a été invité à une médiation et elle a demandé
> une période de réflexion supplémentaire .
> ```

English Input

```
Relations between the US and Germany have come under strain
following claims that the NSA bugged Chancellor Angela 's
Merkel 's phone .
```

French (Ground Truth)

```
Les relations entre les États-Unis et l' Allemagne ont été
mises à rude épreuve à la suite de plaintes selon lesquelles
la NSA avait mis sur écoute le téléphone portable de la
chancelière allemande Angela Merkel .
```

LSTM Translation

```
Les relations entre les Etats-Unis et l' Allemagne ont été
mises à rude épreuve suite aux affirmations selon lesquelles
la NSA aurait pris le téléphone de Merkel de la chancelière
Angela .
```

HyperLSTM Translation

```
Les relations entre les États-Unis et l' Allemagne ont été
mises à rude épreuve après que la NSA a attaqué le téléphone
de la chancelière Angela Angela .
```

English Input

```
Germany 's BfV advises executives to consider using simple prepaid
mobiles when on foreign trips because of the risk that smart phones
are compromised .
```

French (Ground Truth)

```
Le BfV d' Allemagne conseille à ses dirigeants d' envisager
d' utiliser de simples téléphones portables prépayés lors de leurs
voyages à l' étranger en raison du risque d' atteinte à
l' intégrité des smartphones .
```

LSTM Translation

```
Le BfV allemand conseille aux dirigeants d' envisager
l' utilisation de mobiles prépayés simples lors de voyages à
l' étranger en raison du risque de compromission des téléphones
intelligents .
```

HyperLSTM Translation

```
Le BfV allemand conseille aux dirigeants d' envisager
l' utilisation de téléphones mobiles prépayés simples lors de
voyages à l' étranger en raison du risque que les téléphones
intelligents soient compromis .
```

English Input

```
I was on the mid-evening news that same evening , and on TV the
following day as well .
```

French (Ground Truth)

```
Le soir-même , je suis au 20h , le lendemain aussi je suis à la
télé .
```

LSTM Translation

```
J' étais au milieu de l' actualité le soir même , et à la
télévision le lendemain également .
```

HyperLSTM Translation

```
J' étais au milieu de la soirée ce soir-là et à la télévision le
lendemain .
```